\documentclass[sigconf]{acmart}
\usepackage{supertabular}
\usepackage{graphicx}

\AtBeginDocument{%
  }

\copyrightyear{2025}

\acmYear{2025}

\setcopyright{rightsretained}

\acmConference[KDD '25] {Proceedings of the 1st Workshop on ``AI for Supply Chain: Today and Future" @ 31st ACM SIGKDD Conference on Knowledge Discovery and Data Mining V.2}{August 3, 2025}{Toronto, ON, Canada.}

\acmBooktitle{Proceedings of the 1st Workshop on "AI for Supply Chain: Today and Future" @ 31st ACM SIGKDD Conference on Knowledge Discovery and Data Mining V.2 (KDD '25), August 3, 2025, Toronto, ON, Canada}

\acmISBN{979-8-4007-1454-2/25/08}

\acmDOI{10.1145/XXXXXX.XXXXXX}

\begin{document}

\title{Measuring Time Series Forecast Stability for Demand Planning}

\author{Steven Klee}
\authornote{Both authors contributed equally to this research.}
\email{sklee@amazon.com}
\affiliation{%
  \institution{Amazon Web Services}
  \city{Seattle}
  \state{WA}
  \country{USA}
}

\author{Yuntian Xia}
\authornotemark[1]
\email{alicixia@amazon.com}
\affiliation{%
  \institution{Amazon Web Services}
  \city{Atlanta}
  \state{GA}
  \country{USA}
}

\renewcommand{\shortauthors}{Klee, Xia}

\begin{abstract}
Time series forecasting is a critical first step in generating demand plans for supply chains. Experiments on time series models typically focus on demonstrating improvements in forecast accuracy over existing/baseline solutions, quantified according to some accuracy metric. There is no doubt that forecast accuracy is important; however in production systems, demand planners often value consistency and stability over incremental accuracy improvements. Assuming that the inputs have not changed significantly, forecasts that vary drastically from one planning cycle to the next require high amounts of human intervention, which frustrates demand planners and can even cause them to lose trust in ML forecasting models. We study \emph{model-induced stochasticity}, which quantifies the variance of a set of forecasts produced by a single model when the set of inputs is fixed. Models with lower variance are more stable.

Recently the forecasting community has seen significant advances in forecast accuracy through the development of deep machine learning models for time series forecasting. We perform a case study measuring the stability and accuracy of state-of-the-art forecasting models (Chronos, DeepAR, PatchTST, Temporal Fusion Transformer, TiDE, and the AutoGluon best quality ensemble)  on public data sets from the M5 competition and Favorita grocery sales. We show that ensemble models improve stability without significantly deteriorating (or even improving) forecast accuracy. While these results may not be surprising, the main point of this paper is to propose the need for further study of forecast stability for models that are being deployed in production systems. 
\end{abstract}

\received{23 May 2025}
\received[revised]{11 June 2025}
\received[accepted]{11 June 2025}

\maketitle

\section{Introduction}
Forecasting future product demand is the first step in a supply chain system. Planners require accurate forecasts so that they can meet customer demand by delivering the right quantity of products to the right locations at the right time.  When designing a forecasting system, it is common to perform backtesting experiments, measuring the performance of different models across test forecast windows.  For a chosen forecast start date, training and cross validation are done over dates before the start date, and testing is done over the window after the start date. One or more metrics may then be reported on the test window(s) so that the performance of different models can be compared.

The simplicity of reporting a single global accuracy metric is convenient for comparing models, whether that experimentation is done in the context of developing a new model or deciding on which model to deploy in a production system, however it does not tell the whole story. One issue comes from simple mathematics:  if multiple time series are forecast over a multi-week horizon, then the same or similar values of a global metric (for example root mean squared error) can be achieved by infinitely many different combinations of individual forecasts. 

Many local statistical models (e.g., ARIMA) are deterministic, meaning the same set of inputs will yield the same set of forecast outputs as long as the hyperparameters (number of lags, number of differences, order of moving average) are kept constant. However this is not always the case, especially for many of the new state-of-the-art deep learning models. Potential sources of what we will call \emph{model-induced stochasticity} include randomized train/test split methods or stochastic gradient descent methods in the optimization engine. Coupled with non-convex loss functions, this means the same set of forecast inputs can yield different outputs.

In practice, variance in demand forecasts presents a significant challenge to demand planners. A signal whose global accuracy is acceptable but highly variable frustrates planners who ultimately make purchasing decisions across different planning horizons based on different vendor lead times. We propose therefore that it is important to consider model stability when deploying forecasting models in production systems. 

We quantify forecast stability by using the same model with the same hyperparameter settings to perform training and inference on a fixed data set. By repeating this process several times, changing only the random seed, we obtain multiple forecasts drawn from some unknown sample space, and we compute the variance of the outputs over all items to be forecasted and across the forecast time horizon. Precise details are outlined in Section \ref{sec:methodology}.

We note also that there is a second dimension to this problem, namely cycle-to-cycle change, which measures the extent to which model outputs change from one forecast snapshot date to the next. For the purposes of this paper we fix the set of inputs so that the output variance can be attributed to the model alone. Cycle-to-cycle change is undoubtedly important and worthy of additional study. 

The fundamental question we wish to address in this paper is: how much inherent stochasticity do different models introduce; or conversely, how stable are different forecasting models? We use the AutoGluon-TimeSeries library \cite{agtimeseries-paper, autogluon-documentation} to train multiple models on publicly-available data sets that represent different forms of consumer demand: the M5 data set \cite{m5-paper, m5-data}, which contains Walmart product demand in North America; and the Favorita data set \cite{favorita-data}, which contains grocery demand for a South American retail chain.

For each data set and each forecasting model, we measure forecast stability through the coefficient of variation (CV) across ten forecast runs with different random seeds but otherwise the same sets of inputs and model hyperparameters. We then demonstrate trade-off between overall model accuracy and forecast stability. Our main results demonstrate that, perhaps unsurprisingly, ensembled models are more stable than individual deep learning models while also achieving comparable or better accuracy.  

The rest of the paper is structured as follows. In Section \ref{sec:methodology}, we provide an overview of our methodology. In Section \ref{sec:results} we summarize our results on forecast stability and forecast accuracy. Details on hyperparameters and the AutoGluon ensemble model are included in the Appendices.

\section{Methodology} \label{sec:methodology}

\subsection{Measuring model stability}

Given $N$ samples drawn from a distribution, the coefficient of variation is defined as $CV = \frac{\sigma}{\mu}$, where $\sigma$ and $\mu$ are, respectively, the sample standard deviation and mean. The $CV$ is a normalized measure of dispersion of the sample about the mean. When the sample is a set of model outputs, lower values of $CV$ correspond to higher model stability.

In the domain of time series forecasting, suppose we have $M$ time series with histories of length $T$ and values $\{x_{i,t}\}_{i=1}^{M}{\ }_{t=1}^T$. A forecasting model predicts future values $\{\hat{y}_{i,t}\}_{i=1}^M{\ }_{t=T+1}^{T+H}$, where $H$ is the length of the forecast horizon. If a model outputs a distributional forecast, we take $\hat{y}_{i,t}$ to be the center (mean) of the forecast distribution. Re-training and performing inference $R$ times yields multiple forecasts $\{\hat{y}_{i,t,r}\}_{i=1}^M{\ }_{t=T+1}^{T+H}{\ }_{r=1}^R$. Treating the  set of forecasted values for each time series and timestamp as a sample drawn from an unknown distribution gives $M \cdot H$ different coefficients of variation $CV(i,t)$, computed over the sample of $R$ different forecasts $\{\hat{y}_{i,t,r}\}_{r=1}^R$. 

The $CV$ metric has a few drawbacks. After stating the drawbacks we will describe how they can be mitigated in the context of demand forecasting.

\begin{enumerate}
    \item If $\mu$ is negative, then so is $CV$, which does not make sense as a measure of dispersion.
    \item If $\mu$ is zero, $CV$ is undefined.
    \item The $CV$ value can be skewed if $\mu$ and $\sigma$ are small. For example, if $\sigma = 10^{-3}$ and $\mu = 10^{-8}$, then $CV = 10^5$.
\end{enumerate}

Most time series forecasting libraries do not safeguard against negative forecasts because there may be use cases where negative forecasts are valid (e.g., weather forecasts). Even in the context of demand forecasting, small or zero forecast values may be expected, especially when working with intermittent/sparse demand. To overcome these issues, we apply the following forecast post-processing logic before computing $CV(i,t)$: 

\begin{enumerate}
    \item Replace negative forecasted values with zeros. This is a common approach in demand forecasting to overcome the issue of negative forecasted values. 
    \item Round each forecasted value to the nearest integer. This overcomes the issue of small values of $\mu$ which can arise when the values of $\hat{y}_{i,t,r}$ are near zero; and once again we argue it is sensible in the context of forecasting physical units of demand, which must be integers. 
    \item If $\mu = 0$ and $\sigma = 0$, declare $CV = 0$. This happens only in the case of constant zero forecasts. If $\sigma = 0$ it makes sense to also say $CV = 0$ independent of the mean because the data are unchanging. 
\end{enumerate}

\begin{figure}
    \centering
    \includegraphics[width=\linewidth]{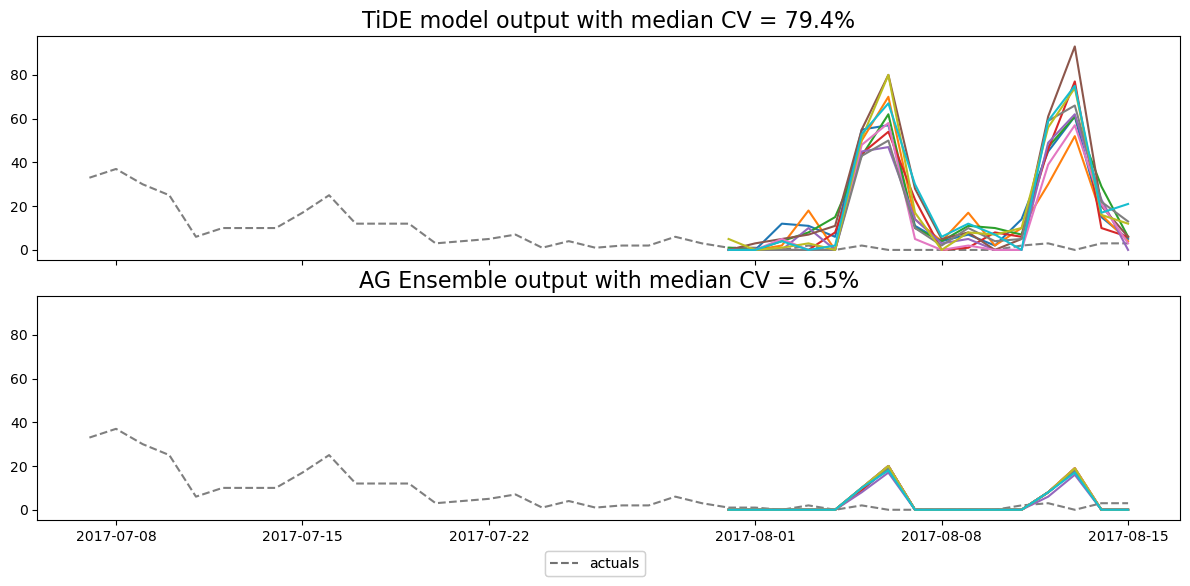}
    \caption{Model outputs across multiple runs for an item from the Favorita catalog. Higher $CV$ values (top) indicate a less stable forecast with a wider range of predicted outputs. Lower $CV$ values (bottom) indicate a more stable forecast. Dashed lines are actual values; solid lines are the outputs of 10 forecast runs.}
    \label{fig:cv_example}
\end{figure}

As an example, Figure \ref{fig:cv_example} shows two distribution outputs from an item in the Favorita data set. The TiDE model output has a high $CV$, with the median value of $CV(i,t)$ near $80\%$ as the time variable ranges over the forecast horizon. In comparison, the AutoGluon ensemble produces a more stable forecast, with a median $CV$ value of $6.5\%$. We see this reflected in the plot as the range of values output by the TiDE model have more uncertainty over the $10$ model runs. Again, to reiterate the methodology, each visualization of a run output shown in this plot is the mean forecast value produced by retraining at TiDE/AutoGluon model with the same set of inputs and only changing the random seed. 

\subsection{Measuring forecast accuracy}

In addition to measuring forecast stability we must also demonstrate a trade-off with forecast accuracy. For each data set we measure forecast accuracy using root mean squared error (RMSE) over the test window. For each forecast run, RMSE is computed as 

$$RMSE(r) = \displaystyle \sqrt{\frac{1}{M \cdot H}\sum_{i=1}^M\sum_{t=T+1}^{T+H}\left(\hat{y}_{i,t,r} - x_{i,t} \right)^2}$$

It is important to note that RMSE is measured in the same units of demand as $x$ and $\hat{y}$. Because it is not normalized it cannot be used to make comparisons across data sets with different demand volumes. This could be overcome with proportional metrics such as MAPE, but that leads to a multitude of other issues, especially for intermittent/sparse demand patterns with many zeros. 

\subsection{Summary of data sets}

We use two public data sets of retail demand data in our analysis. 

\begin{itemize}
\item The M5 data set contains sales data for retail products sold by Walmart in North America \cite{m5-data, m5-paper}. It was used in the 2020 M5 accuracy competition hosted through Kaggle. The M5 data are given at a very low level of granularity (product + geographical region). To mitigate some of the effects of data sparsity, we aggregate to the product level. This was a common approach in the M5 contest as well, where aggregation-disaggregation methods were used to forecast lower levels of the hierarchy.
\item The Favorita data set contains sales data for Corporaci\'on Favorita, a large Ecuadorian-based grocery retailer. The data were used in a 2017 forecast accuracy competition, also hosted through Kaggle \cite{favorita-data}. Once again we aggregate to product-level granularity. 
\end{itemize}

The properties of these data are summarized in Table \ref{table:dataset_summary}.

\begin{table*}
\caption{Summary of data sets}
\begin{tabular}{ccccc}
\toprule
Data set & Granularity & $\#$ time series ($M$) & History length ($T$) & Forecast horizon ($H$) \\ 
\midrule
\texttt{M5} & daily & 3049 & 1913 & 28 \\ 
\texttt{Favorita} & daily & 4036 & 1672 & 16 \\
% \texttt{Traffic} & hourly & 862 & 17520 & 24 \\ 
\bottomrule
\end{tabular}
\label{table:dataset_summary}
\end{table*}

\subsection{Summary of models used}

Table \ref{table:model_summary} summarizes the time series forecasting models that will be used for measuring stability and accuracy. This includes some models that do not introduce stochasticity but are nonetheless important for accuracy comparisons. For deep learning models which are used in a standalone fashion, we performed hyperparameter tuning using grid search. Those models are DeepAR \cite{deepar-paper}, Temporal Fusion Transformer (TFT) \cite{tft-paper}, Patch TST \cite{patchtst-paper}, and TiDE \cite{tide-paper}. 

In addition, we include the Chronos model \cite{chronos-paper} as zero shot and with fine tuning, along with the AutoGluon time series ensemble (AG ensemble) \cite{agtimeseries-paper, autogluon-documentation} with ``best quality" presets. The AutoGluon ensemble is a stacked ensemble comprised of $12$ individual models. It includes some of the above deep learning models; however we do not perform hyperparameter tuning for those models, instead relying on the default settings determined by AutoGluon as an AutoML solution. Further details on the ensemble are outlined in Table \ref{table:ag_ensemble_models} of Appendix \ref{appendix:ag_ensemble}.

All model training and execution is done through the AutoGluon TimeSeries Library (AutoGluon release 1.2.0). In Table \ref{table:model_summary} below we indicate whether hyperparameter tuning was performed in the ``HPO" column. The ``is stochastic" column indicates whether the model outputs change based on setting a random seed. The chosen hyperparameter configurations, which were found by grid search, are listed in Table \ref{table:hpo_settings} of Appendix \ref{appendix:hyperparams}.

\begin{table}
\caption{Models used in experiments}
\begin{tabular}{ccc}
\toprule
Model & is stochastic & HPO \\ 
\midrule
Chronos (zero shot) & no & no \\
Chronos (fine tuned) & no & no \\
DeepAR & yes & yes \\
TFT & yes & yes \\
Patch TST & yes & yes \\
TiDE & yes & yes \\
AG ensemble & yes & no \\
\bottomrule
\end{tabular}
\label{table:model_summary}
\end{table}

\subsection{Related work}

Problems around forecast variance have been studied in the work of Godahewa et al.\ \cite{godahewa-et-al} and Pritularga and Kourentzes \cite{pritularga-kourentzes}. Godahewa et al.\ \cite{godahewa-et-al} consider consider multiple perspectives on this. The first, which they call \emph{horizontal} stability, is the variance of a forecast made on a single snapshot date over the time horizon. The second, which they call \emph{vertical} stability, is the variance across a set of forecasts made on different snapshot dates, but for a fixed future timestamp. Pritularga and Kourentzes \cite{pritularga-kourentzes} take a similar perspective, measuring variance in the time domain across multiple forecast snapshot dates. 

In contrast, the focus of this paper is neither horizontal nor vertical stability, but rather model-induced stability/instability. We measure the extent to which a forecasting model can be viewed as a random process where the variance comes only from changing a random seed, when the snapshot date, input data, and forecast horizon are fixed. Godahewa et al.\ \cite{godahewa-et-al} call this \emph{forecast replicability} but do not study this aspect because the focus of their work is mitigation strategies against such forecast variance, which they note can be mitigated through ensembling in this case. This is consistent with our observations, however they do not attempt to quantify forecast replicability, or what we call model-induced stochasticity. 

\section{Results} \label{sec:results}

\begin{figure}
\centering
    \includegraphics[width=\linewidth]{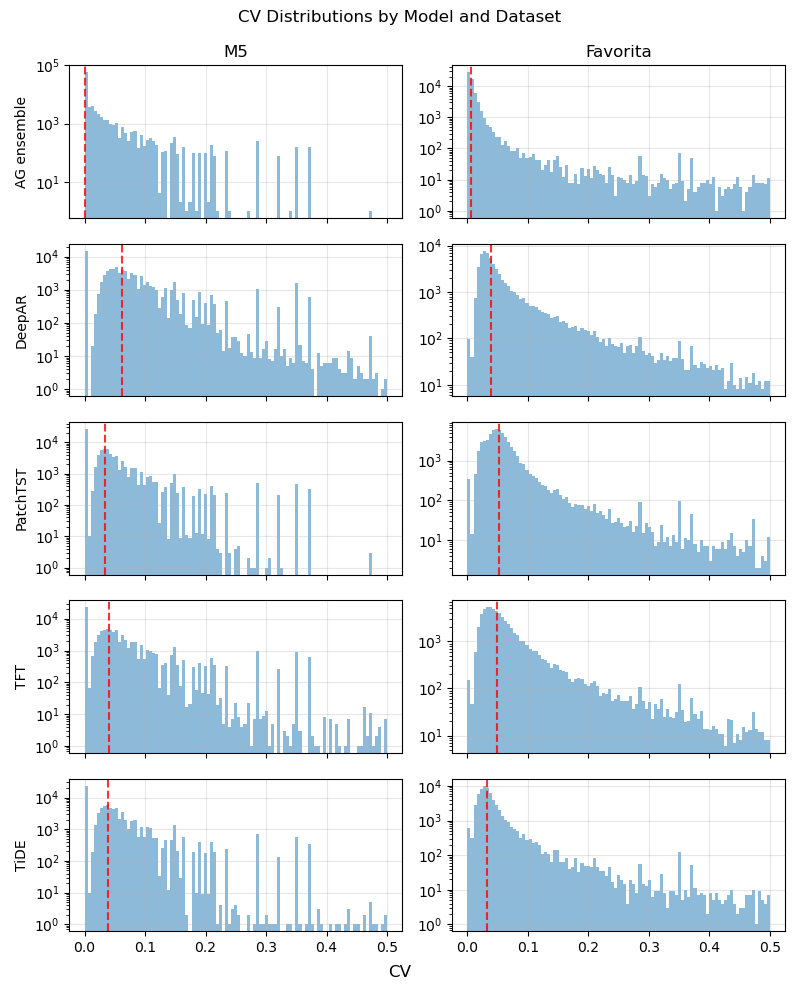}
    \caption{Distribution of CV scores across data sets and stochastic models.}
    \label{fig:cv_distribution}
\end{figure}

Figure \ref{fig:cv_distribution} shows histograms of CV scores over each data set for each stochastic model. The tails of the distributions are clipped for visualization purposes, but as we see in the summary statistics of Table \ref{table:cv_summary_stats} the histograms contain over $90\%$ of the data. Note the vertical axes are on a logarithmic scale. In each plot, the dashed red line shows the median of the distribution.  As noted in Section \ref{sec:methodology}, for each data set we compute $CV$ for $M \cdot H$ different combinations of item and forecast horizon. To simplify the visualizations, we show only the global distributions of all $M \cdot H$ values of $CV$ without separating the time component from the individual items. 

From this we see that the AG ensemble consistently minimizes $CV$, meaning its outputs are more stable than individual deep learning models. This is not surprising given that several of the constituent models in the ensemble (especially Chronos) do not introduce stochasticity. From a business perspective, the following interpretation of the results in Table \ref{table:cv_summary_stats} can be made: at least $10\%$ of the time series forecast by deep learning models can see at least $10\%$, and in some cases nearly $20\%$, normalized variance, \emph{even when trained on the same set of inputs}. In contrast, the AutoGluon ensemble mitigates that variance to less than $5\%$. From a planning perspective, this is a stark difference; $20\%$ changes in output which can be attributed only to stochasticity are likely to be unacceptable for demand planners.

\begin{table}
\caption{Quantiles of the distribution of CV values over data sets and stochastic models}
\begin{tabular}{llrrrr}
\toprule
 Data set & Model & 25\% & 50\% & 75\% & 90\% \\
\midrule
M5 & AG ensemble & 0.000 & 0.000 & 0.012 & 0.048 \\
 & DeepAR & 0.036 & 0.061 & 0.102 & 0.192 \\
 & PatchTST & 0.000 & 0.033 & 0.057 & 0.111 \\
 & TFT & 0.000 & 0.041 & 0.077 & 0.152 \\
 & TiDE & 0.000 & 0.039 & 0.065 & 0.112 \\
\cline{1-6}
Favorita & AG ensemble & 0.007 & 0.011 & 0.022 & 0.043 \\
 & DeepAR & 0.028 & 0.039 & 0.069 & 0.150 \\
 & PatchTST & 0.039 & 0.052 & 0.069 & 0.106 \\
 & TFT & 0.034 & 0.050 & 0.077 & 0.141 \\
 & TiDE & 0.024 & 0.032 & 0.044 & 0.073 \\
\cline{1-6}
\bottomrule
\end{tabular}
\label{table:cv_summary_stats}
\end{table}

\begin{figure}
\centering
    \includegraphics[width=\linewidth]{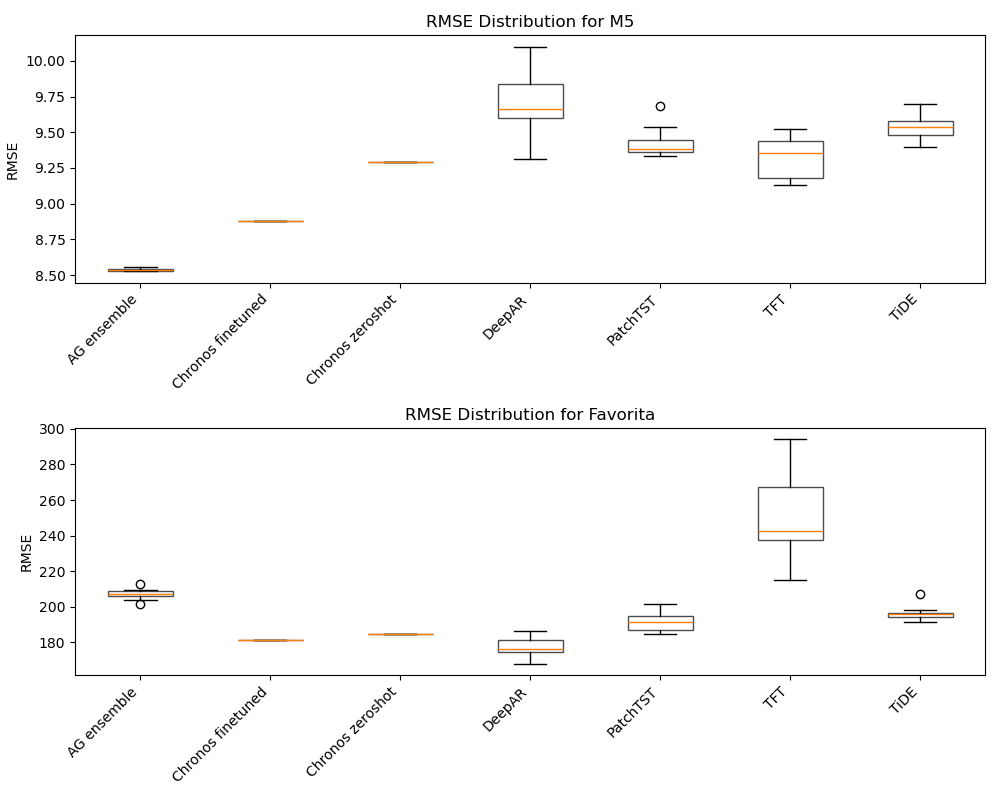}
    \caption{Distribution of forecast error across data sets and models.}
    \label{fig:rmse_distribution}
\end{figure}

Figure \ref{fig:rmse_distribution} shows the distribution of forecast errors as measured by RMSE for each data set and model. For the M5 data we see that the AG ensemble consistently produces more accurate forecasts with tighter distribution of forecast error. The results on Favorita are different. In that case, the AG ensemble is out-performed by both Chronos and the deep learning forecast models. This result is somewhat surprising. On the one hand, the hyperparameter tuning for the deep learning models was not passed into the AG ensemble. Therefore it is conceivable that the tuned models could outperform their counterparts in the AG ensemble. On the other hand, the Chronos models were selected as part of the AG ensemble, and so it is still unexpected that the ensemble would be outperformed by the Chronos models by such a wide margin.  At this point, business planners would need to understand their tolerance for risk and variance in forecast outputs as a trade-off against accuracy improvements. 

\section{Conclusion} 
This study investigated the stability of various time series forecasting models through multiple runs, specifically examining AutoGluon Ensemble, Chronos, TFT, DeepAR, TiDE, and PatchTST. Stability was assessed using both coefficient of variation (CV) of raw outputs and the distribution of Root Mean Square Errors (RMSE).  The results demonstrated that AutoGluon Ensemble exhibited the highest stability among the tested models, showing the lowest variance in both metrics, except the RMSE for the Favorita data set. This superior stability can be attributed to its ensemble approach, as well as the fact that some statistical or otherwise deterministic models were selected in the ensemble. 

The findings suggest that when considering both predictive performance and reliability of results, AutoGluon Ensemble may be particularly suitable for applications where consistent outputs across multiple runs are crucial. Future research could explore how model stability can be impacted by other factors such as model complexity, ensemble configurations, or small perturbations to the inputs. Additionally, as noted in the introduction, a deeper understanding of cycle-to-cycle stability across models is an important practical consideration for forecasting models that are deployed in production systems. It would be interesting to study cycle-to-cycle change in a way that separates the contribution of model-induced stochasticity from changes that can be attributed to the addition of new points in the training data. These insights contribute to our understanding of model reliability in practical applications and can guide practitioners in selecting appropriate forecasting models based on their stability requirements.

\begin{acks}
Thanks to an anonymous referee for pointing out the related works of Godahewa et al. and Pritularga and Kourentzes.
\end{acks}

% \newpage 
\bibliographystyle{ACM-Reference-Format}
\bibliography{main}


\appendix

\section{AutoGluon Ensemble}\label{appendix:ag_ensemble}

Table \ref{table:ag_ensemble_models} shows the component models trained in the AutoGluon ``best quality" ensemble model. The ensemble uses two cross validation windows, and fits a global ensemble by selecting the optimal convex combination of the component model outputs over the validation windows. For more information on individual model configurations we refer to the AutoGluon Model Zoo \cite{autogluon-documentation}.

\begin{table}
\caption{AutoGluon Best Quality Model Ensemble}
\label{table:ag_ensemble_models}
\begin{tabular}{l}
\toprule
Model Name \\
\midrule
Seasonal Naive \\ 
AutoETS \\
Nonparametric Time Series (NPTS) \\ 
Dynamic Optimized Theta \\
Recursive Tabular \\ 
Direct Tabular \\ 
TFT \\ 
Patch TST \\
DeepAR \\
Chronos bolt zero shot \\
Chronos bolt fine tuned \\ 
TiDE \\
\bottomrule 
\end{tabular}
\end{table}

\section{Hyperparameter Tuning}\label{appendix:hyperparams}

Table \ref{table:hpo_settings} shows the hyperparameters learned by grid search for each data set and each of the deep learning forecasting models.
\begin{table*}
\caption{Hyperparameter settings learned by grid search for each data set and deep learning model.}
\label{table:hpo_settings}
\begin{tabular}{@{}llll|llll@{}}
\multicolumn{4}{c}{Favorita} & \multicolumn{4}{c}{M5} \\
\midrule
Model & Hyperparameter & Value & & Model & Hyperparameter & Value \\ 
\midrule
TiDE & dropout\_rate & 0.01 & & TiDE & dropout\_rate & 0.4 & \\
& num\_head & 1 & & & num\_head & 4 & \\
& batch\_size & 64 & & & batch\_size & 256 & \\
& lr & 1e-4 & & & lr & 1e-3 & \\
& distr\_hidden\_dim & 32 & & & distr\_hidden\_dim & 256 & \\
& feat\_proj\_hidden\_dim & 2 & & & feat\_proj\_hidden\_dim & 8 & \\
& encoder\_hidden\_dim & 32 & & & encoder\_hidden\_dim & 256 & \\
& decoder\_hidden\_dim & 32 & & & decoder\_hidden\_dim & 256 & \\
& temporal\_hidden\_dim & 32 & & & temporal\_hidden\_dim & 64 & \\
& num\_layers\_encoder & 1 & & & num\_layers\_encoder & 1 & \\
& num\_layers\_decoder & 1 & & & num\_layers\_decoder & 2 & \\
\cline{1-3} \cline{5-7}
TFT & hidden\_size & 76 & & TFT & hidden\_size & 92 & \\
& dropout\_rate & 0.7 & & & dropout\_rate & 0.2 & \\
& batch\_size & 128 & & & batch\_size & 256 & \\
& num\_head & 1 & & & num\_head & 4 & \\
& lr & 1e-3 & & & lr & 1e-3 & \\
& hidden\_dim & 240 & & & hidden\_dim & 40 & \\
\cline{1-3} \cline{5-7}
PatchTST & num\_encoder\_layers & 4 & & PatchTST & num\_encoder\_layers & 4 & \\
& nhead & 4 & & & nhead & 1 & \\
& batch\_size & 256 & & & batch\_size & 128 & \\
& d\_model & 320 & & & d\_model & 320 & \\
& lr & 1e-4 & & & lr & 1e-4 & \\
\cline{1-3} \cline{5-7}
DeepAR & hidden\_size & 62 & & DeepAR & hidden\_size & 46 & \\
& dropout\_rate & 0.1 & & & dropout\_rate & 0.1 & \\
& batch\_size & 128 & & & batch\_size & 64 & \\
& num\_layer & 3 & & & num\_layer & 4 & \\
& lr & 1e-2 & & & lr & 1e-2 & \\
\midrule
\bottomrule
\end{tabular}
\end{table*}

\end{document}